\documentclass[a4paper,12pt]{article}
\usepackage[english,russian]{babel}
\usepackage[cp1251]{inputenc}
\usepackage[dvipdfm]{graphicx}
\usepackage{mathtext}
\usepackage[T2A]{fontenc}
\usepackage{inputenc,amssymb}
\usepackage{amsfonts}
\usepackage{amssymb}
\usepackage{amsmath}
\usepackage{geometry}
\usepackage{color}

\renewcommand{\le}{\leqslant}
\renewcommand{\ge}{\geqslant}

\renewcommand{\phi}{\varphi}

\newcommand{\eqd}{\stackrel{d}{=}}

\newcommand{\R}{\mathbb{R}}

\newcommand{\eqdef}{\stackrel{\mathrm{def}}{=}}

\title{Extraction of informative statistical features in the problem of forecasting time series generated by It{\^{o}}-type processes}

\author{{Victor Korolev$^{1,2,3,4}$, Mikhail Ivanov$^{1,2,4}$, Tatiana Kukanova$^{1}$,}\\ {Artyom Rukavitsa$^{1}$, Peter Solomonov$^{1}$, Alexander Vakshin$^{1,4}$,}\\ {Alexander Zeifman$^{5,3}$}}

\date{}

\begin{document}

\maketitle

\vspace{-0.5cm}

{\small{$^1$\ Lomonosov Moscow State University, Faculty of Computational Mathematics and Cybernetics}

{$^2$\ Moscow Center for Fundamental and Applied Mathematics}

{$^3$\ Federal Research Center ``Computer Science and Control'', Russian Academy of Sciences}

{$^4$\ Moscow Center for Artificial Intelligence}

{$^5$\ Vologda State University}}







\bigskip

{\bf Abstract.} In this paper, we consider the problem of extraction of most informative features from time series that are regarded as observed values of stochastic processes satisfying the It{\^{o}} stochastic differential equations with unknown random drift and diffusion coefficients. We do not attract any additional information and use only the information contained in the time series as it is. Therefore, as additional features, we use the parameters of statistically adjusted mixture-type models of the observed regularities of the behavior of the time series. Several algorithms of construction of these parameters are discussed. These algorithms are based on statistical reconstruction of the coefficients which, in turn, is based on statistical separation of normal mixtures. We obtain two types of parameters by the techniques of the uniform and non-uniform statistical reconstruction of the coefficients of the underlying It{\^{o}} process. The reconstructed coefficients obtained by uniform techniques do not depend on the current value of the process, while the non-uniform techniques reconstruct the coefficients with the account of their dependence on the value of the process. Actually, the non-uniform techniques used in this paper represent a stochastic analog of the Taylor expansion for the time series. The efficiency of the obtained additional features is compared by using them in the autoregressive algorithms of prediction of time series. In order to obtain pure conclusion that is not affected by unwanted factors, say, related to a special choice of the architecture of the neural network prediction methods, we used only simple autoregressive algorithms. We show that the use of additional statistical features improves the prediction.
\smallskip


\section{Introduction}

Let $X_1,X_2,\ldots$ be a time series that consists of the values of a stochastic process $X(t)$ that are observed at equidistant time moments $t_1,t_2,\ldots$ so that $X_i=X(t_i)$, $i\ge1$. The principal assumption is that the process $X(t)$ satisfies the {\it It{\^{o}} stochastic differential equation}
\begin{equation}\label{Ito}
dX(t)=a(t)dt+b(t)dW(t).
\end{equation}
Here $W(t)$ is the standard Wiener process and the coefficients $a(t)$ and $b(t)$ are random and, in general, unknown.

Stochastic processes given by \eqref{Ito} are considered in many fields of science, first of all, in physics and financial mathematics. In financial mathematics, special versions of \eqref{Ito} are popular, in particular, the geometric Brownian motion
\begin{equation}\label{GBM}
dX(t)=aX(t)dt+b X(t)dW,
\end{equation}
with $a\in\mathbb{R}$, $b>0$, is the basic model describing the evolution of stock prices, financial indexes, etc. There are many generalizations of \eqref{GBM} with special forms of dependence of $a$ and $b$ on $X(t)$ and other random processes, see, e. g., \cite{CoxIngersollRoss1985, Leland1985, Heston1993, BarlesSoner1998, Shiryaev1998}, and stochastic volatility models \cite{HullWhite1987, DermanKani1994, Dupire1994}. Unlike these papers, here we assume that the drift $a(t)$ and diffusion $b(t)$ coefficients are assumed to be unknown and random.

The choice of equation \eqref{Ito} as a model of real data can have the following grounds. In information theory, if there is no prior information concerning the nature of statistical regularities in data, it is customary to follow the maximum entropy principle \cite{Jaynes1957a, Jaynes1957b}.
This principle is well known and widely used in modern methods of machine learning and theory and practice of neural networks \cite{Haykin2001}. According to this principle, if the set of possible states of a complex system is known, but the probability distribution of these states is known only up to some average (integral) characteristics, then the optimal mathematical model of the probability distribution on the set of states should have maximum (differential) entropy (e. g., see \cite{ShoreJohnson1980, Jaynes1982, HarremoeTopsoe2003}). So, if $f(x)$ is the unknown probability density of a random variable $X$ and the values
\begin{equation}\label{1}
\mu_i=\int\nolimits_{-\infty}^{\infty}g_i(x)f(x)dx,\ \ \ \ i=0,1,\ldots,m,
\end{equation}
are given, where $m\in\mathbb{N}$ and $g_i(x)$, $i=1,\ldots,m$, are known functions, then the best model for $f$ is the probability density $f_0$ that delivers maximum to the differential entropy
$$
H(f)=-\int\nolimits_{-\infty}^{\infty}f(x)\log f(x)dx,
$$
under conditions \eqref{1}. It is not difficult to verify that the optimal model $f_0(x)$ has the form
\begin{equation}\label{2}
f_0(x)=\exp\Big\{-1+\sum_{i=0}^m\lambda_ig_i(x)\Big\},
\end{equation}
where the coefficients $\lambda_i$ are determined by the system \eqref{1}.

In particular, if $g_0(x)\equiv 1$, $g_1(x)\equiv x$, $g_2(x)\equiv x^2$, $g_i(x)\equiv 0$, $i\geqslant 3$, then $f_0(x)$ is the densidy of the normal distribution. These conditions are quite understandable: the density should be supported by the whole real line $\mathbb{R}$ (that is, the probability that the random variable $X$ falls into any interval with non-zero length is positive) and the mathematical expectation and variance of $X$ must be finite. Thus, the marginal distributions of the increments of processes of the form \eqref{Ito} with non-random and constant coefficients $a(t)\equiv a$ and $b(t)\equiv b$ have maximum entropy among stochastic processes with continuous sample paths, since the increments of the Wiener process have normal distributions. This means that if the process $X(t)$ evolved in an ideal energy- or information-isolated system, that is, if there was no influence of the environment, then Wiener processes would have to be the best models of the process under consideration. However, almost no real systems (e. g., financial, ecological, physical) can be assumed to be energy- or information-isolated. The deviations of the empirical distributions from the normal (heavy tails, leptokurtosity, assymetry, etc.) often observed in practice are caused by that the influence of the unpredictable environment makes the coefficients $a(t)$ and $b(t)$ random, so that the increments of $X(t)$ have scale-location-mixed normal distributions. It is well known that the non-trivial mixed normal distributions always have heavier tails than the pure normal law (see, e. g., \cite{GnedenkoKorolev1996} or \cite{KShSh2024}). Furthermore, the process $X(t)$ can be non-stationary with the coefficients $a(t)$ and $b(t)$ varying in time in an unpredictable and irregular way. So, model \eqref{Ito} is very general and does not seem unrealistic. If the coefficients $a(t)$ and $b(t)$ of this model are reliably reconstructed from the observed time series, then it is possible to obtain additional latent information concerning the process under consideration. It is also possible to use this information in order to construct more accurate forecasts of the future behavior of the process under consideration.

Here we do not impose any additional constraints like stationarity or homogeneity that are often considered for the sake of convenience of statistical inference. In practice, these conditions can seriously narrow the set of admissible models so that no model in this set will adequately describe the evolution of $X(t)$.

The aim of this paper is to determine, what additional statistical information can be extracted from a time series in order to make the predictive algorithms more accurate, and how these most informative features can be extracted from the time series under consideration. We consider a single univariate time series and restrict ourselves to use only the statistical information that is contained in this time series. In some sense, we undertake an effort to pull ourselves by hair out of the data swamp. We impose no additional conditions like stationarity, self-similarity, etc. The only very general assumption is that the time series is regarded as the observed values of stochastic processes satisfying the It{\^{o}} stochastic differential equations with unknown random drift and diffusion coefficients.

We do not attract any additional information and use only the information contained in the time series as it is. Therefore, as additional features, we use the parameters of statistically adjusted mixture-type models of the observed regularities of the behavior of the time series. Several algorithms of construction of these parameters are discussed. These algorithms are based on statistical reconstruction of the coefficients which, in turn, is based on statistical separation of normal mixtures. We obtain two types of parameters by the techniques of the uniform and non-uniform statistical reconstruction of the coefficients of the underlying It{\^{o}} process. The reconstructed coefficients obtained by uniform techniques do not depend on the current value of the process, while the non-uniform techniques reconstruct the coefficients with the account of their dependence on the value of the process. Actually, the non-uniform techniques used in this paper represent a stochastic analog of the Taylor expansion for the time series.

We discuss some methods of statistical separation of mixtures aiming at the problem of uniform reconstruction of the It{\^{o}} process \eqref{Ito} in Section 2. Here we also describe possible ways of reconstruction of the coefficients of the It{\^{o}} representation of the observed process. The methods of statistical separation of mixtures including weighted EM-algorithm, weighted $\ell_2$ minimization, the method of weighted empirical distribution functions are described in Section 3. In Section 4 we discuss several ways of enriching the training sample (in other words, the feature space) by the parameters of statistical regularities observed while the window slides along the time axis. In Section 5 the use of alternative kernels (Student and logistic) is discussed. The non-uniform technique of state-dependent reconstruction of the coefficients of the It{\^{o}} process is presented in Section 6. In Section 7 we present the results of computational experiments and compare and discuss the accuracy and computational efficiency of the prediction methods constructed by combining the techniques mentioned above. The purpose of the present research is determination of the optimal combinations of prediction techniques that use the (numerical or functional) parameters of statistical regularities of the time series under consideration as additional features. The efficiency of the obtained additional features and the corresponding statistical techniques is compared by the accuracy of the autoregressive algorithms of prediction of time series that use these additional features and techniques. To avoid possible distortions of the resulting conclusions concerning the quality of predictive algorithms, say, caused by a special choice of the architecture of the neural networks, we used only simple autoregressive algorithms.

The methods considered below are purely statistical. However, since these methods are based on statistical separation of mixtures on sliding windows, the volume of available information is always restricted by the window width, it is impossible to speak of such asymptotic statistical properties of the proposed estimators and predictors as consistency, asymptotic normality, efficiency, etc. The quality of estimators and predictors is compared by traditional accuracy measures such as MAE, RMSE and percentage of correctly predicted directions (DIR).

In what follows the symbol $\mathcal{L}[Y]$ will denote the distribution law of a random variable (or random vector) $Y$.

\section{Extraction of additional uniform features by statistical reconstruction of the coefficients of the It{\^{o}} process}

In this section we consider several methods of statistical reconstruction of the coefficients of the It{\^{o}} equation \eqref{Ito} that are aimed at obtaining estimates of the coefficients averaged over sliding windows. Several approaches to efficient averaging are considered.

\subsection{Method of moving statistical separation of mixtures of probability distributions in the problem of prediction of time series}

In the present paper the coefficients $a(t)$ and $b(t)$ are treated as {\it unknown stochastic processes}. By virtue of the random character of these functional coefficients, the problem of their reconstruction admits at least two different formalizations: (i) find (random) approximations to the functions $a(t)$ and $b(t)$, that is, find their {\it point estimates}; and (ii) find a statistical estimate of the {\it probability distributions} of the random variables $a(t)$ and $b(t)$. In the second case, if the analytic properties of these coefficients, for example, the form of their functional dependence on the process $X(t)$ (as it is assumed in \cite{CoxIngersollRoss1985, Leland1985, Heston1993, BarlesSoner1998}) are known, then it is possible to find the estimates of the numeric parameters of these models.

First consider the problem (ii). Let $N\ge1$ and $t_0=0<t_1<\ldots<t_N$ be the time instants at which the process $X(t)$ is observed. For simplicity assume that $t_i-t_{i-1}=1$ for any $i\ge1$. Recall that $X_i=X(t_i)$, $i=1,\ldots,N$.

Since the increments of the Wiener process $W(t)$ have normal distributions, the first step in solving this problem is replacement of the stochastic differential equation \eqref{Ito} by the stochastic {\it difference} equation
\begin{equation}\label{SDE}
X(t_i)-X(t_{i-1})=a(t_{i})+b(t_i)W_i,
\end{equation}
where $\{W_i\}_{i\ge 1}$ are independent random variables with the same standard normal distribution.

From \eqref{SDE} it follows that the distribution of an increment $X_i-X_{i-1}$ of the process $X(t)$ can be approximated by the normal mixture of the form
\begin{equation}\label{inc}
{\sf P}\big(X_i-X_{i-1}<x\big)\approx {\sf E}\Phi\Big(\frac{x-A_i}{B_i}\Big),\ \ \ x\in\mathbb{R},
\end{equation}
where $\Phi(x)$ is the standard normal distribution function,
$$
\Phi(x)=\frac{1}{\sqrt{2\pi}}\int\limits_{-\infty}^{\infty}e^{-y^2/2}dy,\ \ \ x\in\mathbb{R},
$$
$A_i\in\mathbb{R}$ and $B_i>0$ are some random variables. In turn, the distribution of the random variables $A_i$ and $B_i$ (with respect to which the expectation in \eqref{inc} is taken) can be approximated by a discrete distribution so that, instead of \eqref{inc}, a finite normal mixture can be used as approximation to the distribution of the increment $X_i-X_{i-1}$:
\begin{equation}\label{finitemixture}
{\sf P}\big(X_i-X_{i-1}<x\big)\approx \sum\limits_{k=1}^Kp_k\Phi\Big(\frac{x-a_k}{b_k}\Big)\eqdef F(x),\ \ \ x\in\mathbb{R},
\end{equation}
where $K\in\mathbb{N}$, $a_k\in\mathbb{R}$, $b_k>0$, $p_k\ge0$, $k=1,\ldots,K$, $p_1+\ldots+p_K=1$. Some estimates of accuracy of approximation \eqref{finitemixture} can be found, say, in \cite{Korolevetal2015}.

Obviously, the parameters $p_k$, $a_k$ and $b_k$ depend on $i$ as well and can vary with $t_i$ changed by $t_{i+1}$. The choice of $K$ is up to the researcher: small $K$ worsen the accuracy, large $K$ worsen the performance. The value of $K$ can be determined by some information criteria (AIC, BIC, etc.).

Statistical regularities of the behavior of the processes $X(t)$, $a(t)$, $b(t)$ vary in time (in general, in an irregular way). Therefore, in general, it is extremely unlikely to expect that there exists a universal mixing distribution in \eqref{inc}. Therefore, to analyze the changes of statistical regularities in the behavior of the process under consideration, the problem of statistical estimation of the parameters of the mixture on the right-hand side of \eqref{finitemixture} (i. e., the problem of statistical separation of mixture \eqref{finitemixture}) should be successively solved on time intervals (``windows'') moving along the time axis. This means that the mixture parameters (shift (drift) parameters $a_k$, scale (diffusion) parameters $b_k$ and the weights $p_k$ of the components) are estimated as functions of time. In the book \cite{Korolev2011} this procedure was described as a tool for the decomposition of the volatility of stochastic process and was called {\it moving separation of mixtures} (MSM). As this is so, the length of the moving time interval (``window width'') should be large enough to ensure an appropriate accuracy of the resulting estimates and at the same time it should not be too large to avoid ``oversmoothing'' of the resulting estimates. The MSM techniques demonstrated high efficiency in time series analysis in various fields \cite{BBGKKS2019, GK2019, GK2021, VG2021, GK2022, GV2022, KorolevXu2024}.

From what has been said it follows that in the procedure of practical reconstruction of the coefficients $a(t)$ and $b(t)$ of \eqref{Ito} the main role is played by the method of statistical separation of finite mixtures of normal distributions. The properties of the corresponding numerical procedure such as performance, stability, reliability, accuracy are critically important because within the MSM techniques this procedure must be executed repeatedly on sliding windows.

\subsection{Separation of mixtures by maximization of likelihood with EM-algorithm}

If there is no reliable prior information concerning the structure of the process $X(t)$, then the problem of determination of the coefficients $a(t)$ and $b(t)$, that is, the problem of their statistical reconstruction, becomes most important for the retrospective analysis, practical interpretation and/or prediction of its future behavior. Various aspects of this problem were touched upon in many papers, e. g., see \cite{Yoshida1992, Florens-Zmirou1993, Genon-CatalotJacod1993, Genon-CatalotJacod1994, LamourouxLehnertz2009, WeiShu2016} and the references therein. However, in these papers the coefficients of equation (1) are assumed to be {\it known} functions of time, possibly also depending on the process $X(t)$ itself and some unknown numeric or vector parameters.

In practice, the conventional approach to statistical separation of mixtures assumes the application of the maximum likelihood method for the construction of estimates for the mixture parameters $p_k$, $a_k$ and $b_k$, $k=1,\ldots,K$. Within the first approach, on the $i$th window the estimates of the parameters $p_k^{(i)}$, $a_k^{(i)}$ and $b_k^{(i)}$, $k=1,\ldots,K$ are sought as the values of $p_k$, $a_k$ and $b_k$, $k=1,\ldots,K$ that deliver maximum to the log-likelihood function
$$
L(p_k, a_k, b_k,\, k=\overline{1,K}\,| X_{i-n+1},\ldots,X_i)=\sum\limits_{j=i-n+1}^{i}\log\bigg(
\sum\limits_{k=1}^K\frac{p_k}{b_k}\varphi\Big(\frac{X_j-a_k}{b_k}\Big)\bigg)
$$
under the conditions $p_k\ge0$, $k=1,\ldots,K$, and $p_1+\ldots+p_K=1$.

Since the likelihood function corresponding to mixture \eqref{finitemixture} is rather cumbersome, does not have a global maximum and in some cases tends to infinity, if the scale parameter of at least one component tends to zero \cite{RednerWalker1984}, the maximum likelihood estimates can be found only numerically. The conventional numerical procedure used for this purpose is iterative EM-algorithm (E=Expectation, M=Maximization). The general properties of the EM-algorithm are described in \cite{Dempster1977, McLachlanKrishnan1997}. Although the EM-algorithm is proximal (each next iteration increases the target likelihood function), it possesses some unfavourable properties: 1) it is greedy (finds the local maximum) \cite{Wu1983}; 2) it is unstable (noticeably depends on the initial approximation and the data, the replacement of only one observation in the sample can considerably change the result) \cite{Wu1983, Korolev2011}; 3) its performance seriously depends on the stopping criterion; 4) it has a tendency to overtraining as applied to finite normal mixtures (sometimes assigns zero values to scale parameters), see Chapter 9 in \cite{Bishop2006}. The existence of these properties motivated some authors to conclude that maximization of the likelihood function is an ill-posed problem (see Chapter 9 in \cite{Bishop2006}) and the maximum likelihood principle does not work for normal mixtures \cite{EverittHowell1981}. Therefore it is extremely desirable to have an alternative tool that can provide more confidence in the final results with no less performance.

\subsection{Separation of mixtures by minimization of discrepancy between the empirical distribution function and theoretical mixture}

One of possible alternative procedures consists in using another formalization of the statistical problem. Namely, instead of estimation of the {\it vector parameter} $p_k$, $a_k$ and $b_k$, $k=1,\ldots,K$, consider the problem of estimation of the {\it distribution function} $F(x)$ (see \eqref{finitemixture}) as it is. This problem can be solved on each ($i$th) window by minimizing the discrepancy between $F(x)$ and the empirical distribution function $F_n^{(i)}(x)$ constructed from the observations $X_{i-n+1},X_{i-n+2},\ldots,X_{i}$ over the parameters $p_k$, $a_k$ and $b_k$, $k=1,\ldots,K$.
It should be noted that finite mixtures of normal distributions are identifiable \cite{Teicher1961} and stable with respect to small perturbations: small changes of the parameters $p_k$, $a_k$ and $b_k$, $k=1,\ldots,K$ cause small changes of the distribution function $F(x)$ and vice versa \cite{Korolev2011, Korolevetal2015}. Therefore, from the practical point of view, these two statistical settings are similar.

As concerns the concretization of discrepancy, in \cite{Kolourietal2018} it was proposed to minimize the Wasserstein distance. However, in \cite{QZhangChen2022} it was shown that this method requires considerable computational resources and the resulting estimates are less efficient than those obtained by the EM-algorithm.

In the paper \cite{KarpovKorolevSukhareva2025} it was demonstrated that if the discrepancy between the empirical and theoretical distribution functions is understood as the $\ell_2$-distance, the approach based on the discrepancy minimization is preferable at least in the problem of statistical reconstruction of the coefficients of an It\^{o} stochastic process that requires dynamic separation of finite normal mixtures in the moving window mode.

Let $F_{i,n}(x)$ be the empirical distribution function constructed from the sample $X_{i-n+1},X_{i-n+2},\ldots,X_i$ forming the $i$th window,
$$
F_{i,n}(x)=\frac1n\sum\limits_{j=i-n+1}^{i}\mathbb{I}_{(-\infty,x)}(X_j), \ \ \ x\in\mathbb{R},
$$
where $\mathbb{I}_A(x)$ is the indicator function of a set $A$: $\mathbb{I}_A(x)=1$, if $x\in A$ and $\mathbb{I}_A(x)=0$ otherwise.

As a measure of the distance between between the empirical distribution function $F_{i,n}(x)$ and the theoretical distribution function $F(x)$ (see \eqref{finitemixture}) consider $\ell_2$-distance, a discrete analog of the $L_2$-distance:
$$
\ell_2\big(F_{i,n}(x),\,F(x)\big)=\sum\limits_{j=1}^M\big(F_{i,n}(x_j)-F(x_j)\big)^2,
$$
where $M\in\mathbb{N}$ and $x_1,\ldots,x_M$ are some real points. According to the approach under consideration, statistical separation of mixture \eqref{finitemixture} consists in solving the following constrained optimization problem:
\begin{equation}
\label{L2_task}
    \begin{cases}
       \ {\displaystyle\sum\limits_{j=1}^{M} {\bigg( F_{i,n}(x_j) - \sum\limits_{k=1}^{K} p_k \Phi \Big( \frac{x_j - a_k}{b_k} \Big) \bigg) ^ 2} \longrightarrow \min\limits_{p_k, a_k, b_k,\, k=\overline{1, K}}}\vspace{2mm} \\
       \ p_k \geqslant 0, \, b_k > 0, \, k = \overline{1, K}\vspace{2mm} \\
       \ {\displaystyle\sum\limits_{k=1}^{K} p_k = 1}
    \end{cases}
\end{equation}
This problem \eqref{L2_task} can be numerically solved by various optimization algorithms. For example, in \cite{KarpovKorolevSukhareva2025} the sequential quadratic programming algorithm SLSQP($\ell_2$) was used.

The choice of the number $M$ is an option. In \cite{KarpovKorolevSukhareva2025} it was shown that on each (say, $i$th) window the points $x_1,\ldots,x_M$ should be equal to the order statistics $X_{(i+n-1)}, X_{(i-n+1+h)}, X_{(i-n+1+2h)},\ldots, X_{(i)}$ with equidistant ranks constructed from the sample $X_{n-i+1},X_{n-i+2},\ldots,X_i$.

For the problem \eqref{L2_task}, the performance of the numerical procedures is critical. In \cite{KarpovKorolevSukhareva2025}, a combination of numerical procedures was described that provides (almost) the same value of the likelihood function for the discrepancy minimization approach that is attained by the EM-algorithm, but ensures multiple decrease of the $\ell_2$-distance  between the theoretical mixture and the empirical distribution function, at the same time demonstrating better performance.

Since the maximum likelihood method realized by EM-algorithm and the method minimizing the distance  between the theoretical and empirical distribution functions optimize the parameters of mixture \eqref{finitemixture} with respect to absolutely different criteria, it would be desirable to have a technique that combines these methods. It has already been mentioned that the maximization of the likelihood function in the case of a finite normal mixture is an ill-posed problem. Several regularization techniques were proposed for this problem, e. g., see \cite{OrmoneitTresp1998, Ciupercaetal2003, FraleyRaftery2007, Chenetal2008, Ivanovetal2024} and it was demonstrated that the correspondingly penalized likelihood is not unbounded. In \cite{KarpovKorolevSukhareva2025} it was proposed to use an alternative combined procedure: instead of maximizing the penalized likelihood, it was proposed to minimize ``penalized $\ell_2$-distance '' with the log-likelihood playing the role of the penalty. Actually this procedure is a slight modification of $\ell_2$ minimization problem in which the target function is changed by subtracting the weighted log-likelihood function from the $\ell_2$-functional. The performance of this ``hybrid'' minimization procedure occupies and intermediate position between the ``pure'' $\ell_2$-distance  minimization and the likelihood maximization by the EM-algorithm. As it has been mentioned above, in the case of finite normal mixtures under consideration, this modification of the $\ell_2$-distance minimization problem can be also regarded as a regularization of the maximum likelihood method by replacing the EM iterative procedure by the algorithm of sequential quadratic programming.

\subsection{The methods of uniform reconstruction of the coefficients of the It{\^o} process based on moving separation of mixtures}

The estimates of the parameters of finite normal mixtures obtained by the method of moving separation of mixtures can be used for enrichment of the feature space for training intellectual algorithms of forecasting of time series assumed to be the observed values of an It{\^o} process \eqref{Ito}. There are very many approaches to the construction of additional features in order to use more available information. However, sometimes the recommendations look rather artificial and subjective. In \cite{KorolevXu2024} an approach was used that relies on additional objective information concerning the statistical regularities of the behavior of the observed process. Namely, in \cite{KorolevXu2024} the additional information is represented by the multivariate time series of the estimates of the parameters of finite mixtures obtained by the method of moving separation of mixtures. In fact, the use of additional statistical information while training intellectual predictive algorithms narrows the domain of possible solutions by imposing additional pointers or ``signposts'' and, thus, makes it possible to perform guided training by excluding a priori improbable variants with hopes to make training more efficient and, as a consequence, to make predictions more accurate.

Let $n$ be the window width so that if $i\ge n$, then the $i$th window includes the observations $X_{i-n+1}, X_{i-n+2},\ldots,X_{i}$. Denote the estimates of the parameters $p_k$, $a_k$ and $b_k$ on the $i$th window by $p_k^{(i)}$, $a_k^{(i)}$ and $b_k^{(i)}$, $k=1,\ldots,K$, $i=n,\ldots,N$. Then from \eqref{Ito}, \eqref{inc} and \eqref{finitemixture} it follows that for the joint distributions of $a(t_i)$ and $b(t_i)$ for $i=n,\ldots,N$ we can use the discrete approximations
$$
\mathcal{L}\big[a(t_i),b(t_i)\big]\approx\mathcal{L}\big[\widehat{A}_i,\widehat{B}_i\big],
$$
where $\widehat{A}_i$ and $\widehat{B}_i$ are random variables such that
\begin{equation}\label{abdistr}
{\sf P}\big((\widehat{A}_i,\,\widehat{B}_i)=(a_k^{(i)},b_k^{(i)})\big)=p_k^{(i)},\ \ \ k=1,\ldots,K.
\end{equation}
Relation \eqref{abdistr} provides an approximate solution to problem (ii).

At the same time relation \eqref{abdistr} makes it possible to construct an approximate solution to problem (i). Namely, it is well known that the mathematical expectation of a random variable is the best standard approximation to its value. Hence, given the estimates $a_k^{(i)}, b_k^{(i)}, p_k^{(i)}$ for the parameters of mixture \eqref{finitemixture} on the $i$th window, the best estimate for $a(t_i)$ and $b(t_i)$ is the expectation of distribution \eqref{abdistr} in the sense that
$$
\big(a(t_i),\,b(t_i)\big)\approx\bigg(\overline{a}(t_i)= \sum\limits_{k=1}^Kp_k^{(i)}a_k^{(i)},\, \overline{b}(t_i)=\sum\limits_{k=1}^Kp_k^{(i)}b_k^{(i)}\bigg)=
$$
$$
=\arg\min_{a\in\mathbb{R},b>0}{\sf E}\big((\widehat{A}_i-a)^2+(\widehat{B}_i-b)^2\,\big|\,a_k^{(i)}, b_k^{(i)}, p_k^{(i)}\big),\ \ \ i=n,\ldots,N.
$$
It should be noted that since $a_k^{(i)}, b_k^{(i)}, p_k^{(i)}$ are constructed given a {\it random} sample $X_{i-n+1},\ldots,X_{i}$, the obtained estimates $\overline{a}(t_i)$ and $\overline{b}(t_i)$ are also random so that the two-variate time series $\big(\overline{a}(t_i),\,\overline{b}(t_i)\big)$, $i=n,\ldots,N$, can be regarded as a random reconstruction of the time series
$\big(a(t_i),\,b(t_i)\big)$, $i=n,\ldots,N$, of the values of the coefficients of the process $X(t)$ given by \eqref{Ito}.

Another way of construction of uniform estimates of $\big(a(t_i),\,b(t_i)\big)$, $i=n,\ldots,N$, consists in taking not the conditional expectations of $\widehat{A}_i,\,\widehat{B}_i$ but the conditional medians:
$$
\big(a(t_i),\,b(t_i)\big)\approx\big(\mathrm{med}\widehat{A}_i,\, \mathrm{med}\widehat{B}_i\big)=
$$
$$
=\arg\min_{a\in\mathbb{R},b>0}{\sf E}\big(|\widehat{A}_i-a|+|\widehat{B}_i-b|\,\big|\,a_k^{(i)}, b_k^{(i)}, p_k^{(i)}\big),\ \ \ i=n,\ldots,N.
$$

One more way consists in taking conditional discrete modes
$$
\big(a(t_i),\,b(t_i)\big)\approx\big(a_{k_i}^{(i)},\, b_{k_i}^{(i)}\big),
$$
where the index $k_i$ is defined by the relation $p_{k_i}^{(i)}=\max\big\{p_i^{(i)},\ldots,p_K^{(i)}\big\}$.

It should be noted that in the first case the appropriate measure of the prediction accuracy is RMSE, in the second case it is more reasonable to use MAE while in the third case most reasonable measure of accuracy is the relative number of correctly predicted directions.

\section{Weighted maximization of likelihood and minimization of discrepancy}

The method described above is based on moving separation of finite normal mixtures. As this so, all the observations that fall in the current window are forcedly treated as identically distributed random variables. So, a quite natural question arises: what time moment $t\in[i-n+1,\,i]$ does the obtained reconstructions of the It{\^o} stochastic process actually correspond to? If the window of width $n$ includes observations $X_{i-n+1}, X_{i-n+2}, \ldots, X_{i}$, then the obtained $(3K-1)$-variate vector of estimates of the parameters of mixture $F(x)$ (see \eqref{finitemixture}) and the reconstructions of $a(t)$ and $b(t)$ on an equal footing can be regarded as corresponding to any time moment $t\in[i-n+1,\,i]$, say, to the first of them! In other words, the reconstructed coefficients somehow lag behind the original time series. But for the purpose of forecasting it is required to have adequate estimates without any lags that can be regarded as corresponding to the last available observation.

For this purpose it is possible to take into account the natural ordering of observations $X_1,\ldots, X_N$ in time and modify the methods of solving the problems described above by weighting that assigns greater importance to the last observations as compared to the earlier. This approach can be implemented in several ways that will be presented below.

\subsection{Weighted modification of the method of maximum likelihood}

Most natural and most studied method of assigning weights in the problem under consideration is a corresponding modification of the method of maximum likelihood, see, e. g., \cite{KwonLeeKim2022, Pfeffermann1993, Markatouetal1998, SaegusaWellner2013}. Let $X_{i-n+1}, X_{i-n+2}, \ldots, X_{i}$ are the elements of time series that fall into the $i$th window of size $n$. Let $f(x;\,\theta_{i})$ be the assumed probability density function of each observation, $\theta_{i}$ be the vector of unknown parameters corresponding to the $i$th window. Assume that the the elements $X_{i-n+1},\ldots,X_{i}$ have different ``importance'' or ``plausibility''. Then instead of the classical problem of maximization of (the logarithm of) likelihood
$$
\sum\nolimits_{j=1}^n\log f(X_{i-n+j};\theta_{i})\longrightarrow \max_{\theta}
$$
it seems reasonable to solve the problem
$$
\sum\nolimits_{j=1}^n v_j\log f(X_{i-n+j};\theta_{i})\longrightarrow \max_{\theta},
$$
where $v_1,\ldots,v_n$ are finite nonnegative numbers. For the problem of prediction it is reasonable to assume that $v_j\le v_{j+1}$, $j\in\{1,\ldots,n-1\}$. For example, $v_j=1-p^j$, $j=1,\ldots,n$, for some $p\in[0,1)$ (exponential weighting) or $v_j=cj+d$ for some positive $c$ and $d$ (linear weighting). For exponential weighting the value $p=0$ corresponds to the classical likelihood function.

The problem of maximization of the weighted likelihood function can be solved by EM-algorithm. The introduction of weights into step M of EM-algorithm yields the following advantages:

 higher sensitivity to recent observations;

 stability; by diminishing the influence of observations that are distant from the current window, weighting suppresses extra noise and outliers;

 high controllability; by adjusting the weights it is possible to flexibly control the sensibility of the algorithm to old and new data.

\subsection{Weighted modification of the method of minimization of discrepancy}

In the problem of statistical separation of mixture \eqref{finitemixture} by minimizing the $\ell_2$ distance it is possible to assign different weights to the squared differences between the theoretical mixture \eqref{finitemixture} and the empirical distribution function in the points corresponding to order statistics with equidistant ranks, assigning greater weights to the last (in time) observations. If whole sample (window) $X_{i-n+1},\ldots,X_i$ is used to calculate the discrepancy, then the corresponding minimization problem has the form:
$$
W\ell_2\big(F_{i,n}(x),\,F(x)\big)=
\sum_{j=1}^nw_j\bigg[\frac{r_j}{n}-\sum_{k=1}^Kp_k\Phi\Big(\frac{X_{i-n+j}-a_k}{b_k}\Big)\bigg]^2\longrightarrow \min_{p_m, a_m, b_m, m=1,\ldots,K},
$$
where $i$ is the number (position) of the window, $n$ is the window width, $r_j$ is the rank of the element $X_{i-n+j}$ among the order statistics $\{X_{(i-n+1)}\le X_{(i-n+2)}\le,\ldots,\le X_{(i-1)}\le X_{(i)}\}$  ($1=\min\{r_1,\ldots,r_n\},\ n=\max\{r_1,\ldots,r_n\}$),
$w_j$ are the weights calculated, say, in one of the ways described below (see Section 3.4). Note that here the weights may be not normalized so that their sum equals 1.

\subsection{Weighting the empirical distribution function in the problem of minimization of disrepancy}

Instead of the classical empirical distribution function $F_{t,n}(x)$ it is possible to use the function (actually, the statistic)
\begin{equation}
\label{WEDF}
F^*_{t,n}(x)=\sum_{j=t}^{t+n-1}w_{j-t+1}\mathbb{I}(X_j<x),\ \ \ x\in\mathbb{R},
\end{equation}
where $w_1,\ldots,w_n$ are known numbers, $w_i\ge0$ ($i=1,\ldots,n$), $w_1+\ldots+w_n=1$. Within the framework of the problem of forecasting, the weights $w_1,\ldots,w_n$ should be chosen so that they form an increasing sequence.

It should be mentioned that, first, the authors failed to find any references to papers where weighted empirical distribution functions \eqref{WEDF} are considered, that is, this approach can be regarded as new.

Second, the weighted empirical distribution function \eqref{WEDF} inherits some important properties of the classical empirical distribution function. For example,
$$
{\sf E}F^*_{t,n}(x)=F(x);\ \ \ {\sf D}F^*_{t,n}(x)= F(x)\big(1-F(x)\big)\sum_{j=t}^{t+n-1}w^2_{j-t+1},
$$
and, theoretically, if all the sample elements $X_{i-n+1},\ldots,X_i$ have one and the same distribution whatever $n$ is and $\max\{w_1,\ldots,w_n\}\to 0$ as $n\to\infty$, then $F^*_{t,n}(x)\longrightarrow F(x)$ in probability for any $x\in\R$.

Third, this approach demonstrated good performance in the problem of forecasting overlay channel quality, see \cite{Goetal2025}. It is demonstrated in that paper that if sample quantiles calculated from the weighted empirical distribution function \eqref{WEDF} are used in autoregressive prediction algorithms as additional features, then the prediction accuracy increases by $2.6\% \div 25.2$\%
as compared to classical AR and VAR methods (although at the expense of higher computational complexity).

\subsection{Particular ways of weighting}

For the purpose of construction of predictive algorithms it is required that the weights $w_1,\ldots,w_n$ form an increasing sequence. For example, for the weighted empirical distribution function the weights can be chosen as
$$
w_j=\frac{(1-p)(1-p^j)}{n-p(n+1-p^n)}, \ \ \ j=1,\ldots,n,
$$
with some $p\in[0,1)$,(exponential weighting), or
$$
w_j=\frac{2j}{n(n+1)}
$$
(linear weighting). The function $F_{t,n}(x)$ is a special case of $F^*_{t,n}(x)$, where $w_1\!=\!...\!=\!w_n\!=\!\frac1n$ \ ($p=0$ for exponential weighting).

Along with exponential and linear weighting, the following reasoning can be useful.

\smallskip

As is known, in the problem of optimal prediction of the value of $X(t+1)$ from the observations $X(t-n+1),\ldots,X(t-1),X(t)$, under rather general conditions, the arithmetic mean is the best mean square predictor. But in accordance with the general least squares theory (e. g., see \cite{Linnik1958}), when processing observations of {\it unequal accuracy}, the best mean square predictor is defined as as the weighted arithmetic mean in which the weights are proportional to their accuracies (or, more accurately, inversely proportional to their variances). If the original time series was a realization of the process of ordinary Brownian motion $X(t)$, then the variance of $X(t)$ would have been proportional to $t$. Let us in some sense reverse the time and assume that within the problem of prediction of time series the ``accuracy'' of the observation $X(t-i+1)$, is the same as its importance for prediction of $X(t+1)$ and is inversely proportional to some increasing function of $i$. For the Brownian motion, the importance of $X(t-i+1)$ for $X(t+1)$ can be assumed to be the function of the form $ci^{-1}$ with $c>0$.

Since, in general, $X(t)$ may not be a Brownian motion, this function may have another form. In practice, this function can be determined by the following procedure. First, consider all squared one-time-unit increments of the process and calculate their arithmetic mean $s^2_1$. Then, consider all squared two-time-units increments of the process and calculate their arithmetic mean $s^2_2$. Consider all squared three-time-units increments of the process and calculate their arithmetic mean $s^2_3$ and so on to some $m\le N/2$. Approximate the sequence $s^2_i$, $i=1,\ldots,m$, by the function of the form $s^2(i)=ci^{\alpha}$. For this purpose, the parameters $c>0$ and $\alpha>0$ can be estimated by the method of least squares as solutions of the problem
$$
\sum_{i=1}^m(s_i^2-ci^{\alpha})^2\longrightarrow \min_{\alpha,c}
$$
or of the linearized problem
$$
\sum_{i=1}^m(2\log s_i-\log c+\alpha\log i)^2\longrightarrow \min_{\alpha, \log c},
$$
whose solution $\widehat\alpha$, $\widehat c$ can be written out explicitly:
$$
\widehat\alpha=\frac{2\displaystyle{\sum\nolimits_{i=1}^m\log s_i(\log i-m^{-1}\log m!)}}{\displaystyle{m^{-1}(\log m!)^2-\sum\nolimits_{i=1}^m(\log i)^2}},\ \ \ \widehat c=\exp\bigg\{\frac{2}{m}\sum_{i=1}^m\log s_i-\frac{\log m!}{m}\cdot\widehat\alpha\bigg\}.
$$
Then set the weights of observations as
$$
w_i=Ci^{\widehat\alpha},\ \ i=1,\ldots,n,
$$
where the parameter $C>0$ is determined from the condition $w_1+\ldots+w_n=1$, that is,
$$
C=\Big(\sum_{i=1}^ni^{\widehat\alpha}\Big)^{-1}.
$$
Moreover, it is clear that $\widehat c$ does not play any role and can be ignored.

\section{Extension of feature space by the parameters of mixture probability models and procedures of merging of dynamically estimated parameters for the construction of additional time series of features}

In order to use more complete information concerning statistical regularities in the behavior of processes under consideration and to construct more accurate methods and algorithms for prediction of these processes, it is possible to supply the feature space with various additional characteristics. If the feature space is supplemented by the $(3K-1)$-variate time series of the parameters of the mixture-type probability model \eqref{finitemixture} that describes the ``current'' statistical regularities in the behavior of the observed quantities, then the algorithm of construction of the additional time series has crucial importance for the subsequent training to be reliable. The essence of this algorithm is in fixing the rule, according to which the parameters of the finite normal mixture \eqref{finitemixture} (weights, location and scale parameters of the components) estimated on some window, are put in correspondence with the same parameters estimated on the preceding and next windows. This problem arises due to the formal impossibility of identification of the indexes (numbers) of mixture components obtained at different windows because a finite normal mixture is identifiable only up to a permutation of indexes (a finite normal mixture is a weighted sum which does not depend on the ordering of summands). Within this context, some identification algorithms were proposed, see, e. g., \cite{GK2019, Gorsheninetal2020, VG2021, GK2021, GK2022, GV2022}. However, these algorithms either reduced to calculation of empirical or theoretical moments based on the model \eqref{finitemixture}, or had only subjective grounds based on the desirable relative smoothness of the obtained additional time series of the parameters of mixture \eqref{finitemixture}. Here we consider and propose other, possibly more objective, algorithms.

Instead of the $(3K-1)$-variate additional time series of the parameters of the mixture-type probability model \eqref{finitemixture}, the training feature space can be supplemented by the two-variate time series of the reconstructed coefficients $a(t)$ and $b(t)$ of the It{\^o} process \eqref{Ito}. Within this approach both uniform reconstructions described above as well as and non-uniform reconstructions described below (see Section 6) can be used.

Another approach that by-passes the necessity of identification of components, consists in that the additional features are constructed in correspondence with the natural ordering of the values of the distribution function $F(x)$ determined at each window. Namely, as it has already been mentioned, finite normal mixtures \eqref{finitemixture} are identifiable \cite{Teicher1961} in the sense that there is a one-to-one correspondence between the sets of parameters $p_k,\,a_k,\,b_k$, $k=1,\ldots,K$, and the corresponding distribution functions $F(x)$ (up to the permutation of the indexes $k$, it is this circumstance that causes problems of identification of components at different windows). Therefore, instead of transferring the parameters $p_k,\,a_k,\,b_k$, $k=1,\ldots,K$ from one window to another, it is possible to transfer the values of the corresponding distribution functions \eqref{finitemixture}. This can be done in at least two ways.

1. Choose $M\in\mathbb{N}$ and fix the grid $\{x_1,\,x_2,\,\ldots,x_M\}$ on the range of the one-time-unit increments of the process on the whole available time interval. For example, the minimum increment can be taken as $x_1$ and the maximum increment can be taken as $x_M$. The rest points $x_2,\ldots, x_{M-1}$ can be chosen as it was recommended in Theorem 1 of \cite{KarpovKorolevSukhareva2025} (see Section 2.3). Now on each window construct a new $M$-variate vector of the parameters $t_1^{(i)},\ldots,t_M^{(i)}$ by setting
$$
t_j^{(i)}=\sum\limits_{k=1}^{K} p_k^{(i)} \Phi \Big( \frac{x_j - a_k^{(i)}}{b_k^{(i)}} \Big),\ \ \ j=1,\ldots,M,
$$
where $p_k^{(i)},\,a_k^{(i)},\,b_k^{(i)}$, $k=1,\ldots,K$, are the estimates of the parameters of the mixture $F(x)$ obtained from the $i$th window. As this is so, no additional procedures for identifying components on different windows are required.

Obviously, with the exception of $n,\,K,\,M$ and the points $x_1,\ldots,x_M$ that remain constant for any position of the window, all quantities mentioned above depend on the position $i$ of the window. So, along with the original time series $X_1,\ldots,X_N$, the feature space is supplemented by the $M$-variate time series $\big(t_1^{(i)},\,t_2^{(i)},\ldots,t_M^{(i)}\big)$, $i=n,\ldots,N$.

2. Unlike the first method where the values of the arguments of the distribution function $F(x)$ (see \eqref{finitemixture}) remain unchanged as the window slides along the time axis, while the values of $F(x)$ vary, the second method consists in fixing the values of the distribution function $F(x)$ and construction of the set of quantiles of the distribution function $F(x)$, e. g., the set of deciles, on each window.

In fact, the second method can be made non-parametric by replacing the quantiles of $F(x)$ with their empirical analogs, order statistics of the corresponding ranks constructed from the observations that fell into the $i$th window. Here the numbers (ranks) of the order statistics transferred from one window to another must be the same for all windows as well as the width of all windows must be the same. For example, the ranks of the transferred order statistics should form an arithmetic sequence. This choice provides the minimum value of the maximum possible deviation of the empirical distribution function constructed from the observations that fell into the $i$th window, from the theoretical distribution function $F(x)$ in the rest points. The non-parametric version of the second methods is much faster than parametric since there is no need in estimation of the parameters of the mixture \eqref{finitemixture}.

Actually, the idea of both methods is in supplementing the original time series with a multivariate series in which each element is a discrete probability distribution. This approach can be regarded as a development of methods for prediction of histogram time series, see \cite{ArroyoMate2009, Arroyoetal2011, Rakphoetal2021} and the references therein.

\section{Alternative kernels}

\subsection{Student kernels}

As a rule, to reconstruct the coefficients $a(t)$ and $b(t)$ in relation \eqref{Ito}, it suffices to use approximation \eqref{finitemixture} with the number $K$ of components equal to 3, 4 or 5. As this is so, the ``body'' of the theoretical normal mixture corresponding to the distribution of increments of the original time series, is reconstructed satisfactorily. However, the deviation in tail behavior of the fitted mixture and the actual distribution can be noticeable, since the tail probabilities of any finite normal mixture decrease as $O(|x|^{-1}e^{-x^2})$ when $|x|\to\infty$, whereas the tails of the real distribution can be arbitrarily heavy. This deviation can be compensated, if, instead of \eqref{finitemixture}, the approximation
\begin{equation}
\label{Stud}
{\sf P}\big(X_i-X_{i-1}<x\big)\approx \sum_{k=1}^Kp_kT_{r_k}\Big(\frac{x-a_k}{b_k}\Big),\ \ \ x\in\R,
\end{equation}
is used, where $T_{r}(x)$ is the Student distribution function with the shape parameter (``number of degrees of freedom'') $r>0$,
$$
T_{r}(x)=\frac{\Gamma(r+\frac12)}{\sqrt{\pi r}\Gamma(r)}\int_{-\infty}^{x}\Big(1+\frac{y^2}{r}\Big)^{-(r+1/2)}dy,\ \ \ x\in\mathbb{R}.
$$
This approximation is well-justified, since by definition, the Student distribution is a normal scale mixture with the mixing distribution being inverse gamma. So, actually the mixture on the right-hand side of \eqref{Stud} is a scale-location normal mixture in which the mixing distribution is a special mixture of inverse gamma distributions:
$$
{\sf P}\big(X_i-X_{i-1}<x\big)\approx {\sf E}\Phi\Big(\frac{x-A_i}{B_i}\Big)={\sf P}(B_i\cdot\xi+A_i<x),\ \ \ x\in\R,
$$
where $\xi$ is a random variable with the standard normal distribution independent of the random vector $(A_i,\,B_i)$ that has a two-variate distribution function
\begin{equation}
\label{Stud2}
Q_i(x,y)={\sf P}(A_i<x,\,B_i<y)=\sum_{k=1}^{K}p_k{\sf P}\Big(G_{r_k,\,r_k}>\frac{a_k^2}{x^2},\,b_k<y\Big),\ \ \ (x,y)\in\mathbb{R}\times\mathbb{R}^+,
\end{equation}
where $G_{r,r}\eqd \frac1r G_{r,1}$ is a random variable with the gamma-distribution with identical shape and scale parameters equal to $r$. In representation \eqref{Stud2} the parameters $p_k$, $a_k$, $b_k$, $r_k$ depend on the number $i$ (current position) of the window.

Mixture \eqref{Stud} contains more parameters than the normal mixture \eqref{finitemixture} (new $K$ shape parameters $r_k$ of the components appear). Therefore, Student mixture \eqref{Stud} is a more flexible heavy-tailed approximation and is more stable (robust) with respect to the presence of outlying increments of the process. However, as we will see later, these advantages become notable only with window width large enough since the reliable estimation of a larger number of parameters requires larger sample size.

\subsection{Logistic kernels}

By analogy with Sect. 5.1, instead of approximation \eqref{finitemixture}, it is possible to use approximation
\begin{equation}
\label{Log}
{\sf P}\big(X_i-X_{i-1}<x\big)\approx \sum_{k=1}^Kp_kL\Big(\frac{x-a_k}{b_k}\Big),\ \ \ x\in\R,
\end{equation}
where $L(x)$ is the logistic distribution function
$$
L(x)=\frac{1}{1+e^{-x}},\ \ \ x\in\mathbb{R}.
$$
This approximation is also quite reasonable and does not fall beyond the framework under consideration, since as it was shown in \cite{AndrewsMallows1974, Stefanski1990}, the logistic distribution is also a normal scale mixture:
\begin{equation}\label{Normalmixt}
L(x)= \frac{1}{1+e^{-x}}=\int_{0}^{\infty}\Phi\Big(\frac{x}{2z}\Big)dK(z),\ \ \ x\in\mathbb{R},
\end{equation}
where
\begin{equation}\label{Kolm}
K(x)=1-2\sum_{j=1}^{\infty}(-1)^{j+1}\exp\{-j^2x^2\},\ \ \ x\ge 0,
\end{equation}
is the Kolmogorov distribution function that is well known in statistics. The logistic distribution function, as well as its density, is easily computed. Therefore, the logistic distribution is an attractive alternative to the normal distribution. Moreover, with appropriately adjusted location and scale parameters, the logistic distribution is very close to the normal law in the sense of uniform distance ($\rho\le 0.01$) \cite{HillerLiberman2001}. The tails of the logistic distribution function decrease exponentialy, although in considerably more slowly than those of the normal distribution.

The density of the logistic distribution has the form
\begin{equation}\label{Ldd}
f(x)=\frac{e^{-x}}{(1+e^{-x})^2},\ \ \ x\in\mathbb{R}.
\end{equation}

\section{Extraction of additional non-uniform features by statistical reconstruction of the coefficients of the It{\^{o}} process. A stochastic analog of the Taylor formula}

The reconstructions of the coefficients of the It{\^o} process \eqref{Ito} considered above are averaged over possible values of this process within a given sliding window and in this sense are uniform. In this section we will consider an approach to the construction of non-uniform reconstructions of the coefficients $a(t)$ and $b(t)$ with the account of the current value of the time series.

Consider the time series $X_1,X_2,\ldots, X_N$ as a realization of the It{\^o} process
$$
dX(t)=a(t,\,X(t))dt+b(t,\,X(t))dW(t).
$$
Let $n$ be the window width, that is, if $i\geqslant n$, then the $i$th window includes the observations $X_{i-n+1}, X_{i-n+2},\ldots,X_{i}$. In what follows $n$ is assumed to be large enough.

For each window (i. e., for each $i$) denote
$$
m_i=\min\{X_{i-n+1}, X_{i-n+2},\ldots,X_{i}\}, \ \ \  M_i=\max\{X_{i-n+1}, X_{i-n+2},\ldots,X_{i}\}.
$$
Split the interval $[m_i,\,M_i]$ on $J$ nonintersecting equal (or equiprobable) sub-intervals $\Delta_{i,\,1},\ldots,\Delta_{i,\,J}$. To obtain equiprobable sub-intervals, rearrange the sample $X_{i-n+1}, X_{i-n+2},\ldots,X_{i}$ in non-decreasing order and take the order statistics of equidistant ranks as the boundary points of sub-intervals.

Then distribute the observations $X_{i-n+1}, X_{i-n+2},\ldots,X_{i}$ that constitute the $i$th window, over $J$ sub-windows so that the $j$th sub-window contains only those observations from the set $X_{i-n+1}, X_{i-n+2},\ldots,X_{i}$, whose values fall into the sub-interval $\Delta_{i,\,j}$, $j=1,\ldots,J$. Let
$$
\nu_{i,\,j}=\sum_{k=1}^n\mathbb{I}(X_{i-n+k}\in\Delta_{i,\,j}),\ \ \ \nu_i=\min\{\nu_{i,\,1},\ldots,\nu_{i,\,J}\},
$$
that is, $\nu_{i,\,j}$ is the number of observations in the $j$th sub-window, $\nu_i$ is the minimum size of a sub-window of the $i$th window. The numbers $n$ and $J$ should be chosen so that $\nu_i$ is large enough.

Let $j\in\{1,\ldots,J\}$ and $X^{(j)}_{i,\,1},\ldots, X^{(j)}_{i,\,\nu_{i,\,j}}$ be the chronologically ordered observations of the $j$th sub-window. In the initial window $X_{i-n+1}, X_{i-n+2},\ldots,X_{i}$ find observations $\hat X^{(j)}_{i,\,1},\ldots, \hat X^{(j)}_{i,\,\nu_{i,\,j}}$ each of which directly follows after the observations $X^{(j)}_{i,\,1},\ldots, X^{(j)}_{i,\,\nu_{i,\,j}}$. Let
$$
d^{(j)}_{i,\,k}=\hat X^{(j)}_{i,\,k}-X^{(j)}_{i,\,k},\ \ \ k=1,\ldots,\nu_{i,\,j}.
$$
Now fit some distribution function, say, $G_{i,\,j}(x)$, to the sample $d^{(j)}_{i,\,1},\ldots, d^{(j)}_{i,\,\nu_{i,j}}$. For example, $G_{i,\,j}(x)$ can be a discrete distribution of the histogram type, or it can be a finite normal mixture.

A random variable with the distribution function $G_{i,\,j}(x)$ will be denoted $Z_{i,\,j}$. Let
$$
\alpha_{i,\,j}={\sf E}Z_{i,\,j},\ \ \ \beta^2_{i,\,j}={\sf D}Z_{i,\,j}.
$$
If
$$
G_{i,\,j}(x)=\sum_{k=1}^Kp^{(i,j)}_k\Phi\bigg(\frac{x-a^{(i,j)}_k}{b^{(i,j)}_k}\bigg),\ \ \ x\in\R,
$$
then
$$
\alpha_{i,\,j}=\sum_{k=1}^Kp^{(i,j)}_ka^{(i,j)}_k, \ \ \ \beta^2_{i,\,j}=\sum_{k=1}^Kp^{(i,j)}_k\big((a^{(i,j)}_k)^2+(b^{(i,j)}_k)^2\big).
$$
The calculations mentioned above should be performed for all $j=1,\ldots,J$.

Now define the non-uniform reconstruction of the coefficient $a(i,\,X_i)$ as
$$
\overline{a}(i,\,X_i)=\sum_{j=1}^J\alpha_{i,\,j}\mathbb{I}(X_i\in\Delta_{i,\,j}).
$$
In other words, if the value $X_i$ falls into the interval $\Delta_{i,\,j}$, then $a(i,\,X_i)\approx\overline{a}(i,\,X_i)=\alpha_{i,\,j}$.

A simpler reconstruction is based on the non-parametric estimators
\begin{equation}
\label{Aver}
\alpha_{i,\,j}=\frac{1}{\nu_{i,\,j}}\sum_{k=1}^{\nu_{i,\,j}} d^{(j)}_{i,\,k}
\end{equation}

Other ways to choose a predictor consist in calculation of the most probable values
\begin{equation}
\label{Mode}
\alpha_{i,\,j}=\mathrm{mode}Z_{i,\,j}
\end{equation}
or the medians
\begin{equation}
\label{Med}
\alpha_{i,\,j}=\mathrm{med}Z_{i,\,j}.
\end{equation}

In turn, assume that the time series $\{\overline{a}(i,\,X_i),\ i=n+1,\ldots,N\}$ of non-uniform reconstructions of the coefficient $a(t,\,X(t))$ just obtained is a realization of the process $\overline{a}(t)$ also satisfies some It{\^o} equation
\begin{equation}
\label{Itoa}
d\overline{a}(t)=\overline{\overline{a}}\big(t,\,\overline{a}(t)\big)dt+\overline{\overline{b}}\big(t,\,\overline{b}(t)\big)dW(t).
\end{equation}
The procedure described above can be applied to the time series $\{\overline{a}(i,\,X_i),\ i=n+1,\ldots,N\}$ and, as a result we obtain the time series $\{\overline{\overline{a}}\big(i,\,\overline{a}(i,\,X_i)\big),\ i=2n+1,\ldots,N\}$ of the non-uniform reconstructions of the coefficient $\overline{\overline{a}}\big(t,\,\overline{a}(t)\big)$ in \eqref{Itoa}.

The additional time series $\{\overline{a}(i,\,X_i),\ i=n+1,\ldots,N\}$ and $\{\overline{\overline{a}}\big(i,\,\overline{a}(t)\big),\ i=2n+1,\ldots,N\}$ can be regarded as approximations to the reconstructions of the first and second ``discrete'' derivatives of the original process $X(t)$. Therefore, in order to forecast the process $X(t)$ it is possible to use the autoregressive relation
\begin{equation}
\label{TAR2}
X_{i+1}=\sum_{k=0}^{p-1}c^{(i)}_{1,k}X_{i-k}+c^{(i)}_{2,k}\overline{a}(i-k,\,X_{i-k})+
c^{(i)}_{3,k}\overline{\overline{a}}\big(i-k,\,\overline{a}(i-k,\,X_{i-k})\big)+\varepsilon_{i+1},
\end{equation}
where $p\in\mathbb{N}$ is the order of autoregression and the coefficients $c^{(i)}_{m,k}$ are determined anew on each window by the (weighted) least squares techniques.

For the sake of better performance, instead of \eqref{TAR2} it is possible to use the first-order autoregression
\begin{equation}
\label{TAR1}
X_{i+1}=\sum_{k=0}^{p-1}c^{(i)}_{1,k}X_{i-k}+c^{(i)}_{2,k}\overline{a}(i-k,\,X_{i-k})+\varepsilon_{i+1}.
\end{equation}

\section{The results of computational experiments}

The combinations of the features and corresponding techniques of their extraction were applied to the prediction of the time series of measurements of magnetic flux density. We considered the problem of reconstruction of the coefficients of the It{\^o} process fitted as a model of the evolution in time of the projection of the magnetic field (magnetic flux density) $B_x$ onto the coordinate axis $x$ connecting the centers of the Sun and the Earth. The data was registered by the Global Geospace Science (GGS) Wind apparatus (the spacecraft placed in the Lagrange point between the Earth and the Sun). The data is accessible via the web-interface https://omniweb.gsfc.nasa.gov/form/om\_filt\_min.html of the NASA database omniweb.gsf.nasa.gov. We considered the data registered from 00:00:00 1 January, 2020, to 23:59:00 31 December, 2020. The time lag between observations was equal to 1 minute that is convenient for tracing the details in the evolution of the interplanetary magnetic field. The original time series is depicted on Figure 1.

\renewcommand{\figurename}{{Fig.}}


Along with the original time series $B_x$, two more its slightly smoothed versions were analysed. The first smoothed time series was constructed in the following way. The elements of the original time series were grouped in sequential non-intersecting pairs and the half-sum of the elements of each pair was taken as an element of the new, smoothed, time series. The second smoothed version of $B_x$ was constructed in a similar way by averaging each four elements of the original series. The purpose of this smoothing is to get rid of high-frequency noise.

The predictive properties of practically all combinations of the statistical characteristics mentioned above and techniques of their extraction from the original time series were considered and tested.

When optimization algorithms (such as classical and weighted versions of the EM-algorithm or the minimization of the $\ell_2$-discrepancy) in the sliding window mode were used, the final values of the mixture parameters obtained on a window were taken as the initial approximation for these algorithms on the next window.

Non-uniform reconstruction of the coefficients of the It{\^o} representation was used to compute the predicted value of the time series in accordance with the autoregressive scheme \eqref{TAR1}.

Computations were performed on the supercomputer ``Lomonosov'' at the Moscow State University.

The metrics characterizing the difference between the predicted and true values of the time series were considered as the quality measures of additional features and the techniques of their extraction. Predictions one step ahead were analysed. Three metrics were computed: MAE, RMSE and DIR (the percentage of correctly predicted directions). The corresponding results are presented in Tables 1 -- 9. These tables contain the best values of the MAE, RMSE and DIR metrics obtain on the specified time series with the specified window width, and the corresponding optimal combinations of the additional characteristics and methods of their computation. The following notations are used:

\smallskip

AR($p$) means the classical autoregression of order $p$;

\smallskip

Taylor($p$) means the method of non-uniform reconstruction of the coefficients of the It{\^o} process described in Section 6, $p$ is the order of the corresponding autoregressive scheme ($p=1$ or 2);

\smallskip

Q denotes the equiprobable choice of the sub-intervals of the range of the time series within a window;

\smallskip

U denotes the choice of the sub-intervals with equal length;

\smallskip

$J=m$ means that the number of sub-intervals is equal to $m$;

\smallskip

{\sf Avg} means that the prediction was calculated in accordance with \eqref{Aver}

\smallskip

{\sf Mode} means that the prediction was calculated in accordance with \eqref{Mode}

\smallskip

Norm($K$) means the method of uniform reconstruction of the coefficients of the It{\^o} process based on the mixture model with normal kernels and $K$ components;

\smallskip

Log($K$) means the method of uniform reconstruction of the coefficients of the It{\^o} process based on the mixture with logistic kernels and $K$ components;

\smallskip

Stud($K$) means the method of uniform reconstruction of the coefficients of the It{\^o} process based on the mixture with Student kernels and $K$ components;

\smallskip

WEM(linear weights) means the weighted EM-algorithm with linear weighting;

\smallskip

VAR($p$) means the vector autoregression of order $p$, when the original time series is supplemented by the two time series of the uniformly reconstructed coefficients of the It{\^o} process, and the autoregressive scheme is applied to the corresponding 3-dimensional time series; the accuracy metric is computed only with respect to the original time series;

\smallskip

The results of computational experiment yield the following conclusions.

1. The use of additional statistical features improves the prediction. It becomes more efficient, when the window width is not small providing acceptable accuracy of the corresponding statistical procedures.

2. As a rule, better results are achieved by supplementing the feature space by the coefficients of the It{\^o} process uniformly reconstructed by likelihood maximization with EM-algorithm. The method of $\ell_2$-discrepancy minimization demonstrates slightly worse results. However, it should be noted that, as it was shown in \cite{KarpovKorolevSukhareva2025}, the minimization of $\ell_2$-discrepancy is preferable for the retrospective analysis of time series.

3. As a rule, the use of the estimated quantiles of the mixture or sample quantiles as additional features in autoregressive schemes does not improve the quality of prediction (at least for the time series considered in this paper). However, as it was shown in \cite{Goetal2025}, the use of sample quantiles computed from the exponentially weighted empirical distribution function as additional features demonstrated best results among autoregressive algorithms in the problem of prediction of some quality characteristics of the overlay channel.

4. Weighted procedures improve the quality of prediction. Linear weighting demonstrates best results.

5. The optimal combination of models, features, and methods of their extraction depends on the target quality metric. It is almost impossible to identify the ``most optimal'' combination. The prediction algorithms should be constructed by adaptive choice of the suitable combination of features and techniques.


\renewcommand{\thesection}{}


\vspace{-0.2cm}

\renewcommand{\baselinestretch}{1.1}

\renewcommand{\tablename}{{\rm Table}}

\begin{table}[!ht]
\centering
\caption{Best MAE, original time series $Bx$.}\vspace{3mm}
\begin{tabular}{||p{3cm}|p{2.2cm}|p{8.5cm}||}
\hline
Window width &  Best MAE    & Best algorithm \\
\hline
50           & 0.2999773 & AR(1) \\
100          & 0.2944703 & AR(1) \\
200          & 0.2920770 & AR(1) \\
1080         & 0.2906131 & Taylor(1), Q, $J=9$ \\
2160         & 0.2817369 & Taylor(1), Q, $J=9$ \\
4320         & 0.2696042 & Taylor(1), Q, $J=9$ \\
\hline
\end{tabular}
\end{table}

\begin{table}[!ht]
\centering
\caption{Best RMSE, original time series $Bx$.}\vspace{3mm}
\begin{tabular}{||p{3cm}|p{2.2cm}|p{8.5cm}||}
\hline
Window width & Best RMSE    & Best algorithm \\
\hline
50           & 0.5718501 & AR(1) \\
100          & 0.5591483 & AR(1) \\
200          & 0.5556475 & AR(1) \\
1080         & 0.5507813 & Log(3), WEM(linear weights), VAR(1), {\sf Avg} \\
2160         & 0.5303492 & Log(4), WEM(linear weights), VAR(1), {\sf Avg} \\
4320         & 0.5112293 & Stud(5), WEM(linear weights), VAR(1), {\sf Avg}\\
\hline
\end{tabular}
\end{table}

\begin{table}[!ht]
\centering
\caption{Best DIR, original time series $Bx$.}\vspace{3mm}
\begin{tabular}{||p{3cm}|p{2.2cm}|p{8.5cm}||}
\hline
Window width &  Best DIR     & Best algorithm \\
\hline
50           & 51.1165217 & Taylor(2), U, $J=27$ \\
100          & 51.2009292 & Taylor(2), Q, $J=9$ \\
200          & 51.5442890 & Log(4), WEM(linear weights), VAR(1), {\sf Avg} \\
1080         & 52.2174122 & Stud(5), WEM(linear weights), VAR(3), {\sf Avg} \\
2160         & 52.5200200 & Stud(5), WEM(linear weights), VAR(3), {\sf Avg} \\
4320         & 52.2290305 & Stud(4), WEM(linear weights), VAR(3), {\sf Avg} \\
\hline
\end{tabular}
\end{table}

\begin{table}[!ht]
\centering
\caption{Best MAE, Time series $Bx$ pairwise smoothed.}\vspace{3mm}
\begin{tabular}{||p{3cm}|p{2.2cm}|p{8.5cm}||}
\hline
Window width &  Best MAE    & Best algorithm \\
\hline
100          & 0.3326546 & Taylor(1), Q, $J=4$ \\
540          & 0.3292382 & Taylor(1), Q, $J=4$ \\
1080         & 0.3220519 & AR(1) \\
2160         & 0.3139312 & Taylor(1), Q, $J=13$ \\
\hline
\end{tabular}
\end{table}

\begin{table}[!ht]
\centering
\caption{Best RMSE, Time series $Bx$ pairwise smoothed.}\vspace{3mm}
\begin{tabular}{||p{3cm}|p{2.2cm}|p{8.5cm}||}
\hline
Window width &  Best RMSE    & Best algorithm \\
\hline
100          & 0.5952129 & Taylor(1), Q, $J=50$ \\
540          & 0.5875930 & Log(3), WEM(linear weights), VAR(1), {\sf Avg} \\
1080         & 0.5726880 & Norm(3), WEM(linear weights), VAR(1), {\sf Avg} \\
2160         & 0.5630516 & Log(3), WEM(linear weights), VAR(1), {\sf Avg} \\
\hline
\end{tabular}
\end{table}

\begin{table}[!ht]
\centering
\caption{Best DIR, Time series $Bx$ pairwise smoothed.}\vspace{3mm}
\begin{tabular}{||p{3cm}|p{2.2cm}|p{8.5cm}||}
\hline
Window width &  Best DIR     & Best algorithm \\
\hline
100          & 51.5268065 & Stud(3), WEM(linear weights), VAR(1), {\sf Med} \\
540          & 52.9942280 & Stud(3), WEM(linear weights), VAR(2), {\sf Avg} \\
1080         & 53.1931932 & Norm(3), WEM(linear weights), VAR(3), {\sf Avg} \\
2160         & 53.3850763 & Stud(4), WEM(linear weights), VAR(2), {\sf Avg} \\
\hline
\end{tabular}
\end{table}

\begin{table}[!ht]
\centering
\caption{Best MAE, Time series $Bx$ smoothed over four observations.}\vspace{3mm}
\begin{tabular}{||p{3cm}|p{2.2cm}|p{8.5cm}||}
\hline
Window width &  Best MAE    & Best algorithm \\
\hline
260          & 0.4162940 & Taylor(1), U, $J=6$ \\
520          & 0.4105675 & Taylor(1), U, $J=6$ \\
1080         & 0.4028675 & Taylor(1), U, $J=25$ \\
\hline
\end{tabular}
\end{table}

\begin{table}[!ht]
\centering
\caption{Best RMSE, Time series $Bx$ smoothed over four observations.}\vspace{3mm}
\begin{tabular}{||p{3cm}|p{2.2cm}|p{8.5cm}||}
\hline
Window width &  Best RMSE    & Best algorithm \\
\hline
260          & 0.7028217 & Log(3), WEM(linear weights), VAR(1), {\sf Avg} \\
520          & 0.6924566 & Log(5), WEM(linear weights), VAR(1), {\sf Avg} \\
1080         & 0.6808704 & Log(3), WEM(linear weights), VAR(2), {\sf Avg} \\
\hline
\end{tabular}
\end{table}

\begin{table}[!ht]
\centering
\caption{Best DIR, Time series $Bx$ smoothed over four observations.}\vspace{3mm}
\begin{tabular}{||p{3cm}|p{2.2cm}|p{8.5cm}||}
\hline
Window width &  Best DIR    & Best algorithm \\
\hline
260          & 53.9625360 & Stud(3), WEM(linear weights), VAR(1), {\sf Avg} \\
520          & 54.5209581 & Log(5), WEM(linear weights), VAR(2), {\sf Med} \\
1080         & 54.6459695 & Stud(4), WEM(linear weights), VAR(2), {\sf Avg} \\
\hline
\end{tabular}
\end{table}

\newpage

\renewcommand{\baselinestretch}{0.95}

\renewcommand{\refname}{References}

\end{document}